\documentclass{article}
\usepackage{textgreek}
\usepackage{graphicx} 
\usepackage{amsmath}		% Extended math symbols and environments
\usepackage{amsfonts}		% Extended math fonts
\usepackage{latexsym}		% Extended symbols
\usepackage[authoryear]{natbib}			% The natbib citation package

\usepackage{booktabs}		% Nicer stylings of tables
\usepackage{placeins}		% Enhanced control over floating environments
\usepackage{url}			% Nicer Urls
\usepackage{hyperref}		% Crossrefs in LaTeX
\title{Handwritten Amharic Character Recognition Using a Convolutional Neural Network}
\author{Mesay Samuel Gondere  \\
	Arba Minch University, Faculty of Computing and Software Engineering,\\ (mesay.samuel@amu.edu.et)  \\
	\and 
	Lars Schmidt-Thieme \\
	Information Systems and Machine Learning Lab, 31141 Hildesheim, Germany \\ (schmidt-thieme@ismll.uni-hildesheim.de) \\
	\and 
	Abiot Sinamo Boltena \\
	Ethiopian Institute of Technology-Mekelle, School of Computing \\ (abiotsinamo35@gmail.com) \\
	\and 
	Hadi Samer Jomaa \\
	Information Systems and Machine Learning Lab, 31141 Hildesheim, Germany \\ (hsjomaa@ismll.uni-hildesheim.de) \\
	}

\date{\today}
\begin{document}

\maketitle

\begin{abstract}
Amharic is the official language of the Federal Democratic Republic of Ethiopia. There are lots of historic Amharic and Ethiopic handwritten documents addressing various relevant issues including governance, science, religious, social rules, cultures and art works which are very reach indigenous knowledge. The Amharic language has its own alphabet derived from Ge'ez which is currently the liturgical language in Ethiopia. Handwritten character recognition for non Latin scripts like Amharic is not addressed especially using the advantages of the state of the art techniques. This research work designs for the first time a model for Amharic handwritten character recognition using a convolutional neural network. The dataset was organized from collected sample handwritten documents and data augmentation was applied for machine learning. The model was further enhanced using multi-task learning from the relationships of the characters. Promising results are observed from the later model which can further be applied to word prediction.
\end{abstract}

\section{Introduction}
\label{MesaySamuelsec:1}
% Always give a unique label
% and use \ref{FIRSTAUTHORNAME<label>} for cross-references
% and \cite{<label>} for bibliographic references
% use \sectionmark{}
% to alter or adjust the section heading in the running head

\begin{figure}[tbp]
%\sidecaption
\centering
% Use the relevant command for your figure-insertion program
% to insert the figure file.
% For example, with the graphicx style use
\includegraphics[scale=0.85]{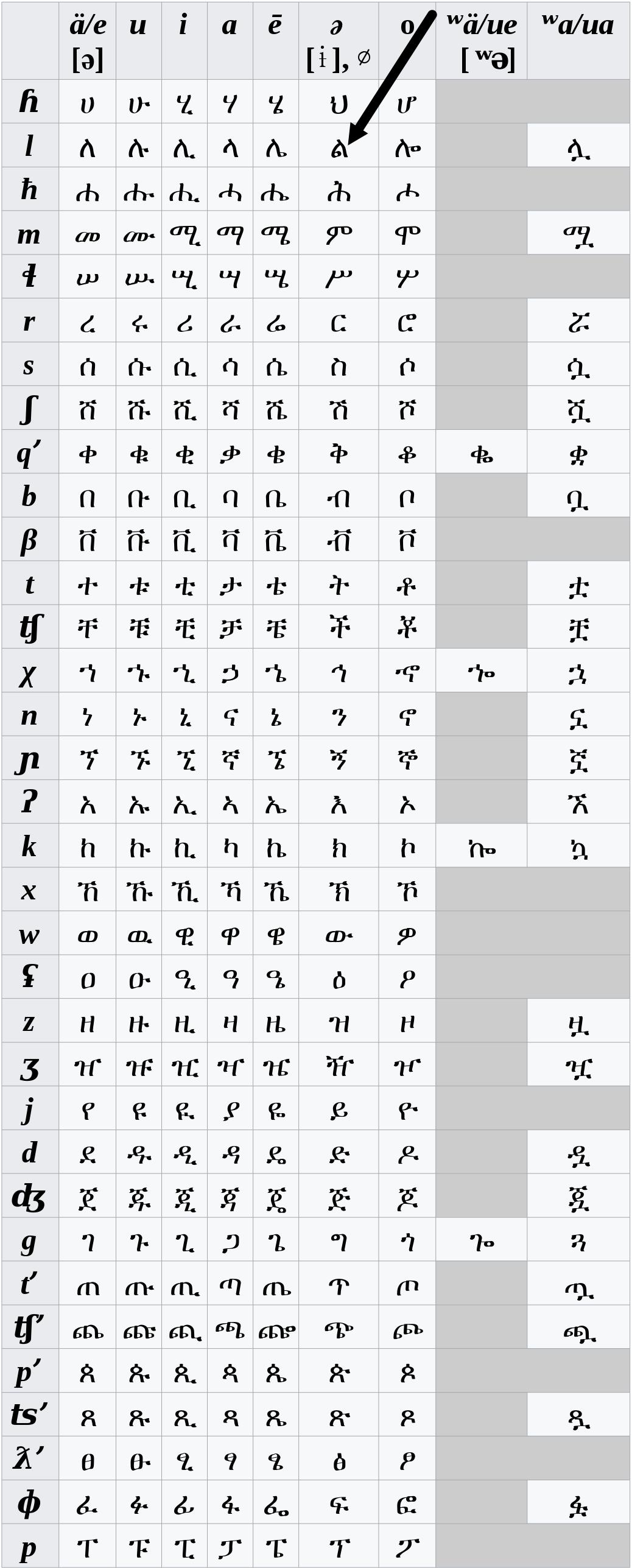}
%
% If no graphics program available, insert a blank space i.e. use
%\picplace{5cm}{2cm} % Give the correct figure height and width in cm
%
\caption{The Amharic Alphabet, e.g.\ The character indicated by the arrow is the 13\textsuperscript{th} label with 2\textsuperscript{nd} row and 6\textsuperscript{th} column.}
\label{MesaySamuel:1}       % Give a unique label
\end{figure}

Amharic language is the official language of the federal government of Ethiopia and other regional states in Ethiopia like Southern Nations, Nationalities, and People Region (SNNPR). It is a Semitic language with its own scripts where other same family languages in Ethiopia share the fonts.  The  Amharic language is believed to be derived from Ge'ez, the liturgical language of Ethiopia. The total number of Ethiopic scripts including Amharic is 446, 20 numerical representations, 9 punctuations, 8 tonal marks, 3 combining marks and 6 special characters, summed to a total 492 numbers of scripts. The Amharic alphabet as shown in Figure ~\ref{MesaySamuel:1} has 265 characters including 27 labialized and 34 base characters with six orders representing derived vocal sounds of the base character \citep{assabie_offline_2011,weldegebriel_deep_2018}. Each character represents a consonant+vowel sequence, but the basic shape of each character is determined by the consonant, which is modified for the vowel. There are lots of historic Amharic and Ethiopic handwritten documents addressing various relevant issues including governance, science, religious, social rules, cultures, and art works which are reach indigenous knowledge. However, these handwritten documents are not available electronically to be accessed and processed by the wider public while getting the advantage of Internet and emerging computing technologies \citep{weldegebriel_deep_2018,meshesha_optical_2007}.

Optical Character Recognition (OCR) is a technology that enables the conversion of different types of written documents, such as scanned paper documents, PDF files or images into editable and searchable data. Basically OCR targets typewritten text, one glyph or character at a time. However, intelligent character recognition (ICR) targets handwritten print-script or cursive text one glyph or character at a time, usually involving machine learning. With regard to Amharic OCR and ICR research, still less has been done and the recognition techniques employed traditional approaches. Research results also show that the performances of the available prototypes are less especially for various quality types of image and different types of fonts. The large number of alphabets, similarity of the characters, unavailability of corpus, and the lack of standard for Amharic fonts are mentioned to be the major reasons that complicated the efforts so far \citep{assabie_offline_2011,meshesha_optical_2007}.

Basically character recognition includes the following phases: pre-processing, segmentation, feature extraction, and classification.  While each of the various stages has impact on the recognition accuracy, the feature extraction technique plays the major role \citep{hirwani_handwritten_2014,purohit_literature_2016}. In this regard, convolutional neural networks (CNN) have the potential to preserve detailed features including the dimensional information. Hence, offline handwritten character recognition systems have achieved successful results with the contribution of convolutional neural networks \citep{xiao_building_2017,pradeep_diagonal_2010}. In handwritten character processing systems, due to some domain artifacts  it is difficult to design a generic system which can process handwritten characters for all kinds of languages \citep{purohit_literature_2016}. Hence, the design of recognition systems for other languages will open a way to look the characteristics of other scripts. 

This paper addresses Amharic handwritten character recognition using convolutional neural networks for the first time. The paper further points out the advantages of multi-task learning which can be implied from the relationships of the Amharic characters. This work explored the recognition pattern of the various tasks related to the Amharic alphabet which will help in addressing the problem and indicating scalability possibilities to other scripts. 

The rest of the paper is organized as follows. Related works are reviewed in the next section. Section ~\ref{MesaySamuelsec:3} outlines the methodology followed for the study and experimental results are discussed under section ~\ref{MesaySamuelsec:4}. Finally, conclusion and future works are forwarded.

\section{Related Works}
\label{MesaySamuelsec:2}

Handwritten character recognition for non-Latin scripts is still an active area of research. While convolutional neural networks are the state of the art techniques applied in most image recognition tasks,  the recent research efforts in handwritten recognition diverse in two categories. The first group emphasizes the improvement of recognition accuracies by trying the possible deeper and complex architectures \citep{elleuch_new_2016,roy_handwritten_2017}. On the contrast there are attempts to simplify the architectures stressing more on the less run time and space complexity of the proposed solutions \citep{xiao_building_2017,zhang_online_2016}. In their literature survey \citet{purohit_literature_2016} outlined the importance of feature extraction techniques in character recognition and revealed the need to address enhancement of algorithms and recognition rates. \citet{rosyda_review_2018} discussed the different methods used to address the challenges of handwritten character recognition and reported CNN to be the best method in terms of getting higher accuracy. As a typical example of the advantages in CNN, \citet{el-sawy_arabic_2017} implemented for Arabic handwritten characters with different parameter optimization methods to increase the performance of CNN. The authors used two CNN layers with 80 and 64 feature maps, two pooling layers and one fully connected layer and reported a  promising result with 94.9\% classification accuracy rate on testing images. 

There are limited research works for Ethiopic character recognition. The possible reasons mentioned are the use of large number characters in the writing, existence of large set of visually similar characters and unavailability of standard dataset until recent attempt \citep{assabie_offline_2011,meshesha_optical_2007}. The first attempt for Amharic offline character recognition was reported by \citet{cowell_amharic_2003}. They approached the problem using template and signature template matching. \citet{assabie_writer-independent_2008} have implemented offline handwritten character recognition for Ethiopic script based on the characteristics of primitive strokes that make up characters. The authors also develop a comprehensive dataset for related research works. A work by \citet{weldegebriel_deep_2018} addressed deep learning for Ethiopian Ge'ez script optical character recognition and  demonstrated the promises of applying convolutional neural networks for Ethiopic scripts. A recent related parallel work by \citet{belay_factored_2019} proposes a CNN based approach for Amharic character image recognition.The later two papers emphasize on frequent Amharic characters and used synthetic printed characters.    

\section{Methodology}
\label{MesaySamuelsec:3}
This section outlines how dataset preparation and techniques of recognition were employed to undertake the study.
\subsection{Dataset Preparation}
\label{MesaySamuelsubsec:31}

\begin{figure}[tbp]
%\sidecaption
\centering
% Use the relevant command for your figure-insertion program
% to insert the figure file.
% For example, with the graphicx style use
\includegraphics[scale=0.85]{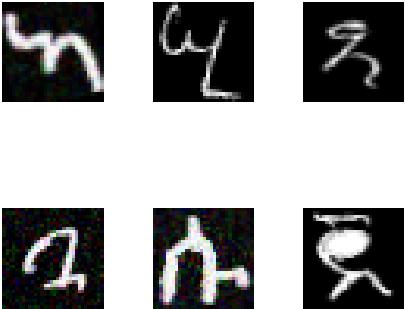}
%
% If no graphics program available, insert a blank space i.e. use
%\picplace{5cm}{2cm} % Give the correct figure height and width in cm
%
\caption{Sample handwritten characters from the dataset}
\label{MesaySamuel:2}       % Give a unique label
\end{figure}

The dataset for this study was organized from the work of \citet{assabie_comprehensive_2009}, a comprehensive Dataset for Ethiopic Handwriting Recognition. A subset of this dataset  contains offline isolated characters freely written by several participants, each participant writing the 265 Amharic alphabets in one page. Twelve unique handwritings were extracted as stratified samples for each 265 Amharic language alphabets from this dataset using python scripts. The ratio 9:2:1 was applied for training, validation and test splits per alphabet. Due to the importance of big dataset in machine learning \citep{wigington_data_2017}, the subsets representing each alphabet were further augmented using data augmentation techniques including -15 to 15 degree random rotations, random noise, and 70-87\% resize (diminish) using python scripts.  Accordingly, 1192500 images (4500 per alphabet), 212000 images (800 per alphabet), and 106000 images (400 per alphabet) were used for the training, validation, and test sets respectively. Finally, the dataset was represented in numpy array format incorporating one-hot encoding for the labels and hence ready for input to the CNN model. Figure ~\ref{MesaySamuel:2} shows sample characters of the dataset.

\subsection{Convolutional Neural Network Architecture}
\label{MesaySamuelsubsec:32}

\begin{figure}[tbp]
%\sidecaption
\centering
% Use the relevant command for your figure-insertion program
% to insert the figure file.
% For example, with the graphicx style use
\includegraphics[scale=0.66]{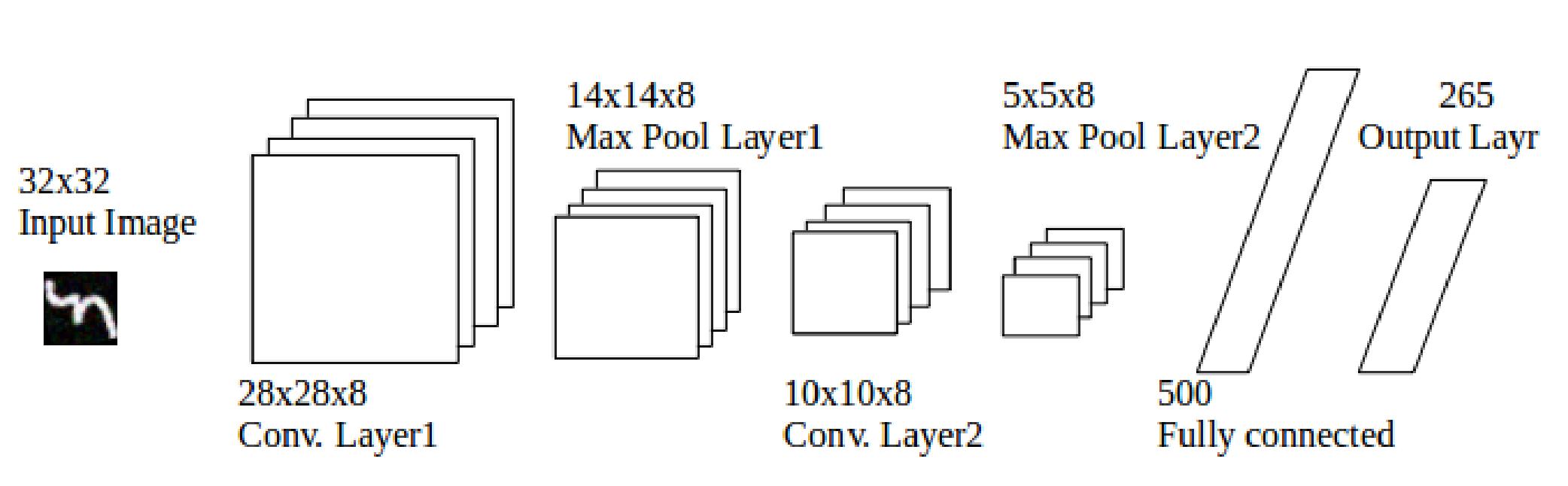}
%
% If no graphics program available, insert a blank space i.e. use
%\picplace{5cm}{2cm} % Give the correct figure height and width in cm
%
\caption{The proposed CNN architecture for Amharic handwritten characters}
\label{MesaySamuel:3}       % Give a unique label
\end{figure}

Convolutional neural network (CNN) is a class of deep neural networks widely used as the state of the art technique in computer vision. CNNs have demonstrated the potential of automatically preserving salient features from the input and hence are not sensitive to variations. The CNN network basically structured as a set of layers including convolution layers, sub-sampling layers and fully connected layer.  From an MxMxC1 input neuron nodes which will be convolved with NxNxC1 filter and stride of one, the convoultion layer outputs (M-N+1)x(M-N+1)xC2. The sub-sampling layers like max-pooling reduce the dimensionality of each feature map while retaining the relevant information. Finally, the fully connected layer or attached with other classifiers outputs the predictions. The rest of the CNN elements including the activation functions and regularizations remain correspondent to any other neural networks for optimization or parameter tuning. This study adapted the CNN architecture from the work of \citet{el-sawy_arabic_2017} which was designed for Arabic handwritten character recognition. Hence, the architecture shown in Figure ~\ref{MesaySamuel:3} was reconstructed through incremental experiments to avoid the encountered high-level over fitting. 

\subsection{Multi-task Learning}
\label{MesaySamuelsubsec:33}

\begin{figure}[tbp]
%\sidecaption
\centering
% Use the relevant command for your figure-insertion program
% to insert the figure file.
% For example, with the graphicx style use
\includegraphics[scale=0.70]{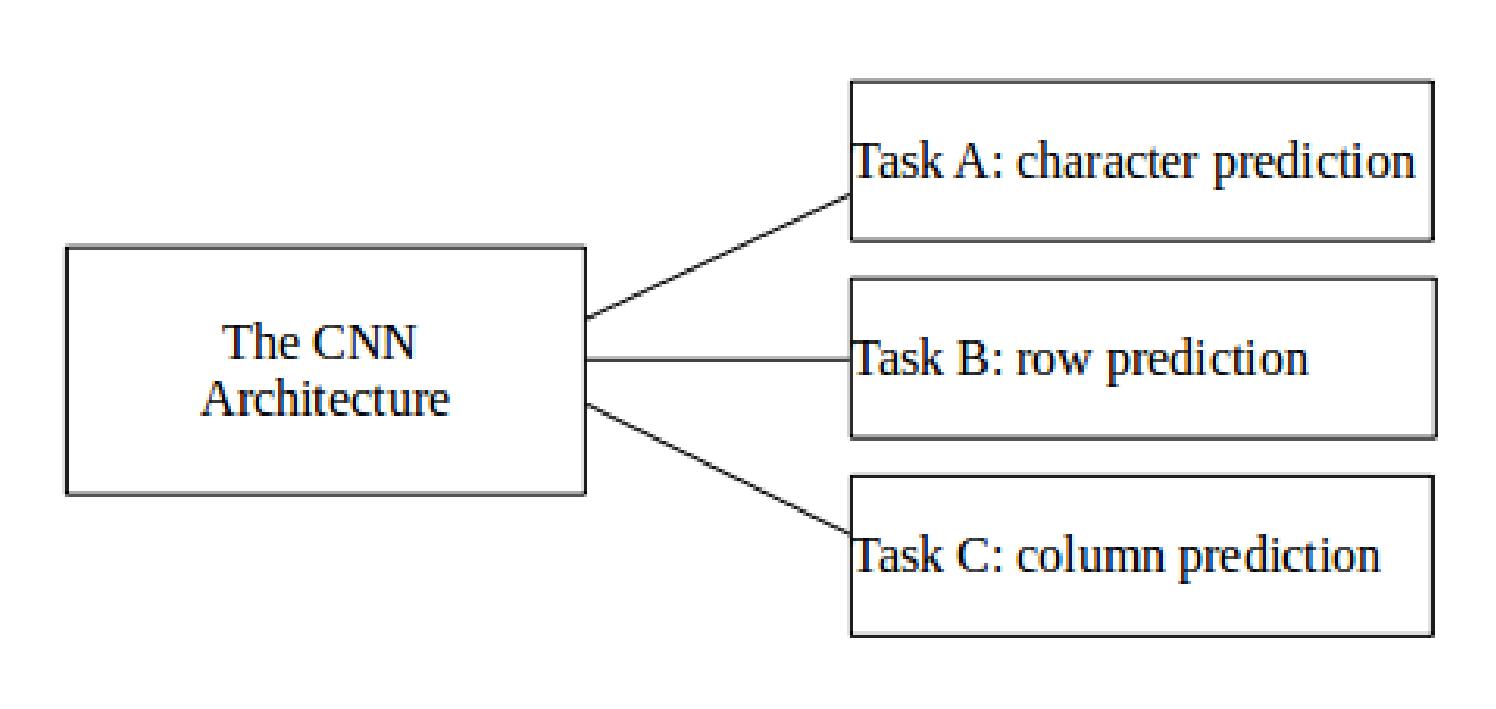}
%
% If no graphics program available, insert a blank space i.e. use
%\picplace{5cm}{2cm} % Give the correct figure height and width in cm
%
\caption{Multi-task learning (hard parameter sharing) using the rows and columns class}
\label{MesaySamuel:4}       % Give a unique label
\end{figure}

Multi-task learning is learning with auxiliary tasks to help improve upon the main task. Technically, it is optimizing more than one loss function in contrast to single-task learning. Multi-task learning has been used successfully across all applications of machine learning as it improves generalization by leveraging the domain-specific information contained in the training signals of related tasks. The widely used multi-task learning approach is hard parameter sharing where the hidden layers between all tasks are shared while keeping several task-specific output layers. Hence the more number of tasks, the more generalization of the main task \citep{ruder_overview_2017}. From the structure of the Amharic alphabet, one can identify two related tasks. These are the row class (1-34) and the column class (1-9) of the alphabet.  Accordingly, for this study these tasks are added by weighting each loss with hard parameter sharing as shown in Equation ~\ref{MesaySamueleq:1} and Figure ~\ref{MesaySamuel:4}. 

\begin{equation}
Loss =\alpha _{1}\cdot l\left( y,\widehat {y}\right) +\alpha _{2}\cdot l\left( y_{2},\widehat { y_{2}}\right) +\alpha _{3}\cdot l\left( y_{3},\widehat { y_{3}}\right)
\label{MesaySamueleq:1}
\end{equation}

\section{Experimental Results}
\label{MesaySamuelsec:4}

All the experiments were performed using TensorFlow on GPU nodes connected to computing cluster at Information Systems and Machine Learning Lab (ISMLL), University of Hildesheim.

\subsection{Convolutional Neural Networks}
\label{MesaySamuelsubsec:41}

The convolutional neural network run with the following hyper parameter settings after repeated incremental experiments: 100 batch size, 0.0001 learning rate, 0.01 L2 regularization, and 0.3 keeping probability for dropout. The choice of the hyper parameters is empirical with a focus on the learning behavior of the model. The experiments were controlled by early stopping when the loss values show no more reduction.

\begin{figure}[tbp]
%\sidecaption
\centering
% Use the relevant command for your figure-insertion program
% to insert the figure file.
% For example, with the graphicx style use
\includegraphics[scale=0.66]{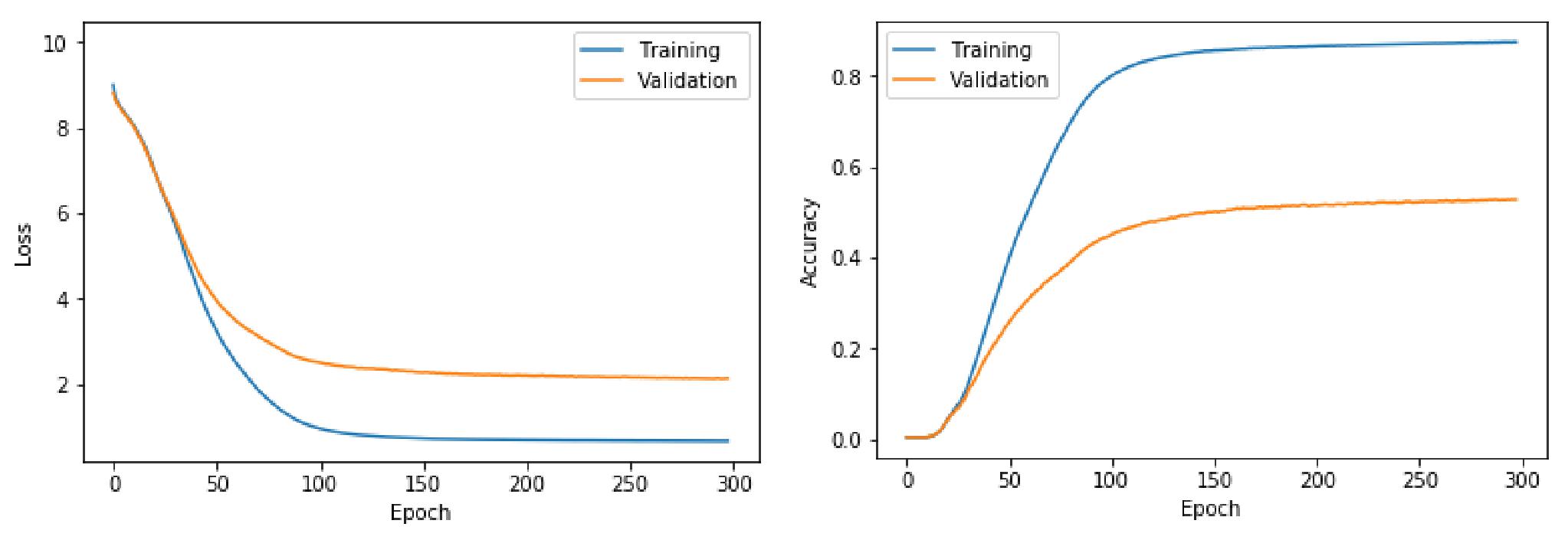}
%
% If no graphics program available, insert a blank space i.e. use
%\picplace{5cm}{2cm} % Give the correct figure height and width in cm
%
\caption{The loss (left) and accuracy (right) curves of the CNN model}
\label{MesaySamuel:5}       % Give a unique label
\end{figure}

As shown in Figure ~\ref{MesaySamuel:5} the model converges after 200 epochs. The loss curve shows a drop from 9.01 to 0.67 and from 8.81 to 2.13 for training and validation sets respectively. On the other hand, 87.48\% and 52.15\% accuracy reached at 300 epochs for training and validation set respectively. Similarly, 2.05 loss and 52.82\% accuracy achieved on test set. This is an average performance attained with CNN without a need for any feature extraction technique.  However, the larger gap between the training accuracy 87.48\% and the test accuracy 52.15\% might be due to the less number of unique characters which is only nine. Even though data augmentation was used to address the varieties in handwriting and helped to this level, it may not encompass the natural varieties to scale unseen data sets.  

\subsection{Multi-task Learning with Different Alpha Values}
\label{MesaySamuelsubsec:42}

\begin{figure}[tbp]
%\sidecaption
\centering
% Use the relevant command for your figure-insertion program
% to insert the figure file.
% For example, with the graphicx style use
\includegraphics[scale=0.66]{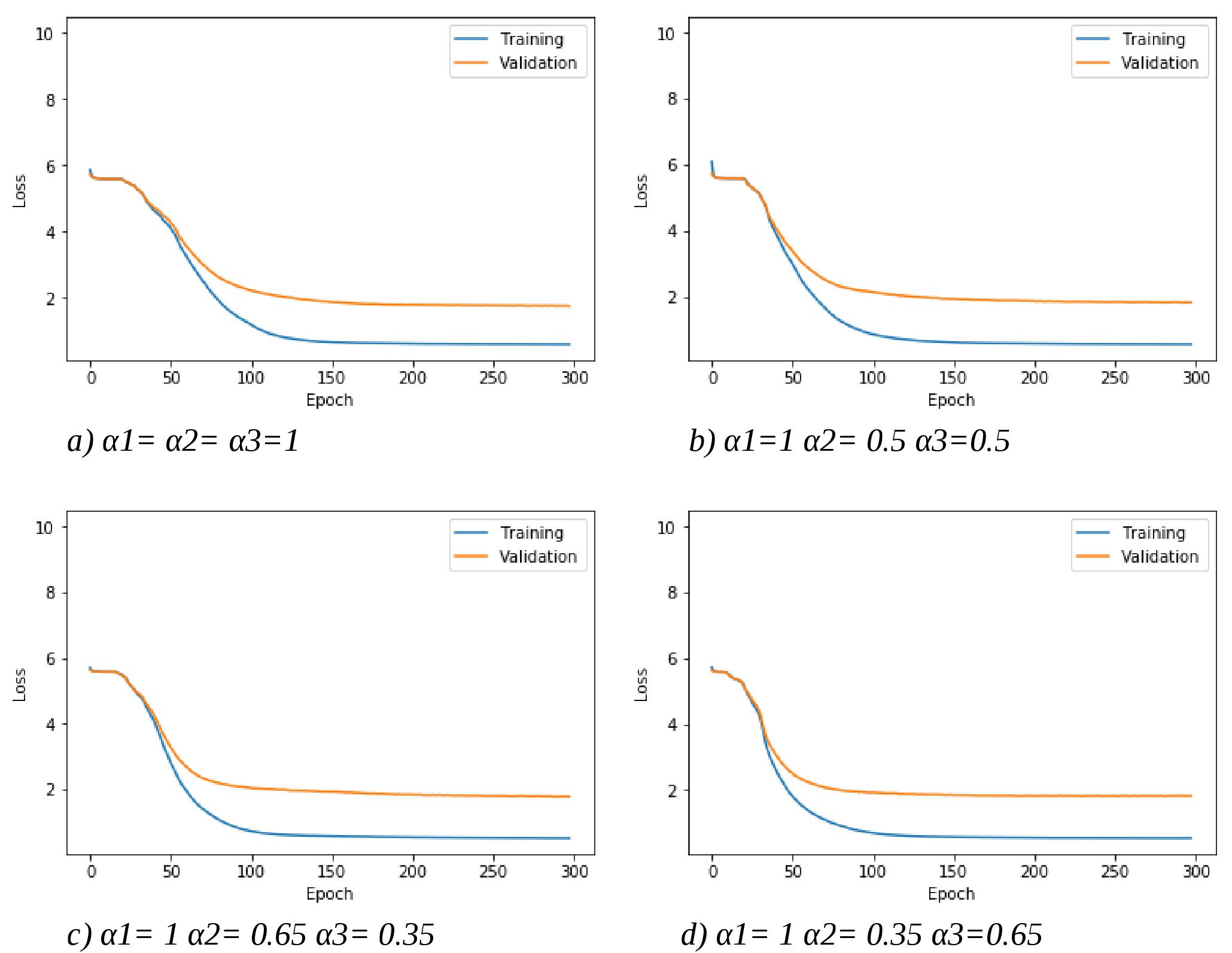}
%
% If no graphics program available, insert a blank space i.e. use
%\picplace{5cm}{2cm} % Give the correct figure height and width in cm
%
\caption{The loss curves for multi-task learning with different alpha values (a-d)}
\label{MesaySamuel:6}       % Give a unique label
\end{figure}

With all similar hyper parameter settings and introduction of multi-task learning in the CNN model, a second phase of the experiment was tested to examine the improvements. Learning the auxiliary tasks (predicting the rows and columns of the alphabets) has helped the learning of the main task (predicting the individual alphabet labels). Different coefficients of alpha values were used in these experiments to identify the degree of influence by each task and the results are shown in Figure ~\ref{MesaySamuel:6} (a-d).

\begin{figure}[tbp]
%\sidecaption
\centering
% Use the relevant command for your figure-insertion program
% to insert the figure file.
% For example, with the graphicx style use
\includegraphics[scale=0.75]{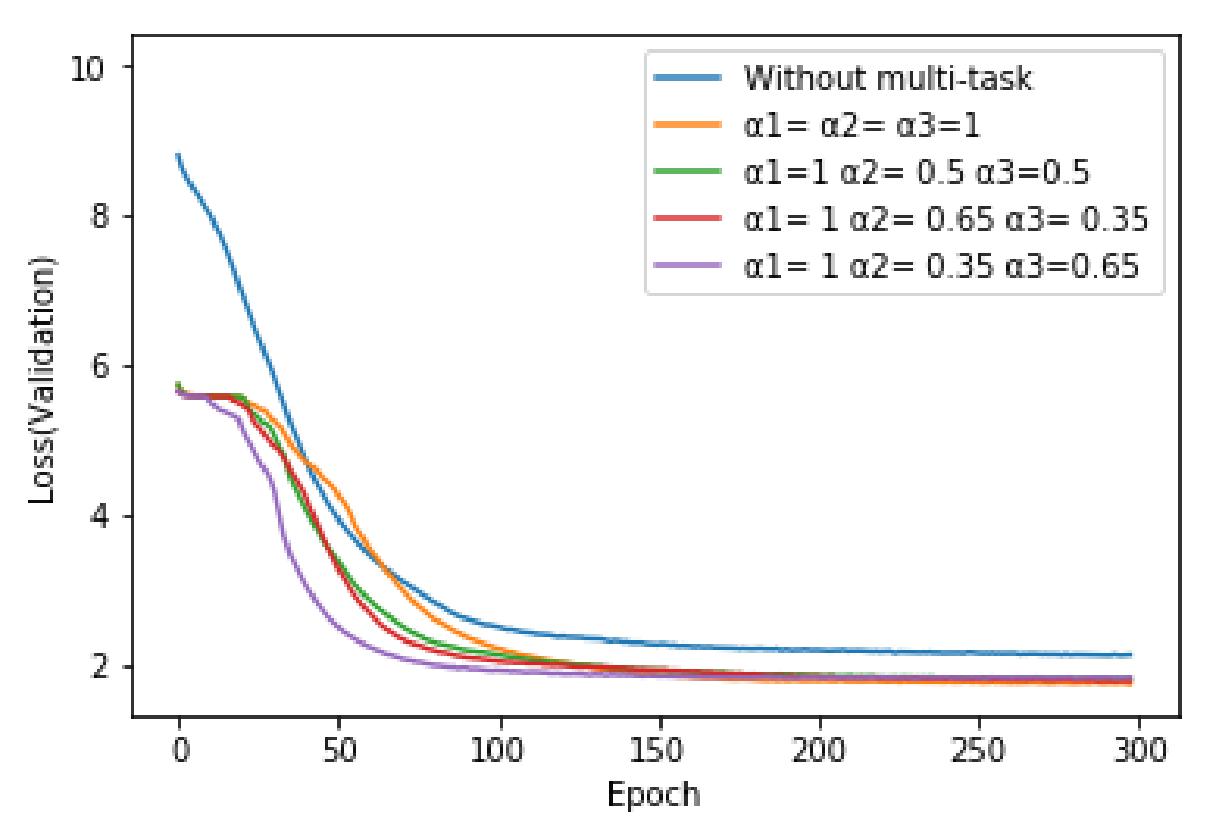}
%
% If no graphics program available, insert a blank space i.e. use
%\picplace{5cm}{2cm} % Give the correct figure height and width in cm
%
\caption{The overall comparison among all the experiments with (a-d) and without
multi-task learning}
\label{MesaySamuel:7}       % Give a unique label
\end{figure}

As it can be easily observed from the above figure a considerable improvement was attained in minimizing the loss with all the multi-task learning experiments. In all the cases the loss drops from [6.10 - 5.70] to [0.58 - 0.52] and [5.74 - 5.64] to [1.83 - 1.74] for training and validation sets respectively. This finding serves as an empirical evidence to show the support of multi-task learning in improving the generalization capability of the model. The overall comparison among all the experiments was put together in Figure ~\ref{MesaySamuel:7}.

\begin{figure}[tbp]
%\sidecaption
\centering
% Use the relevant command for your figure-insertion program
% to insert the figure file.
% For example, with the graphicx style use
\includegraphics[scale=0.66]{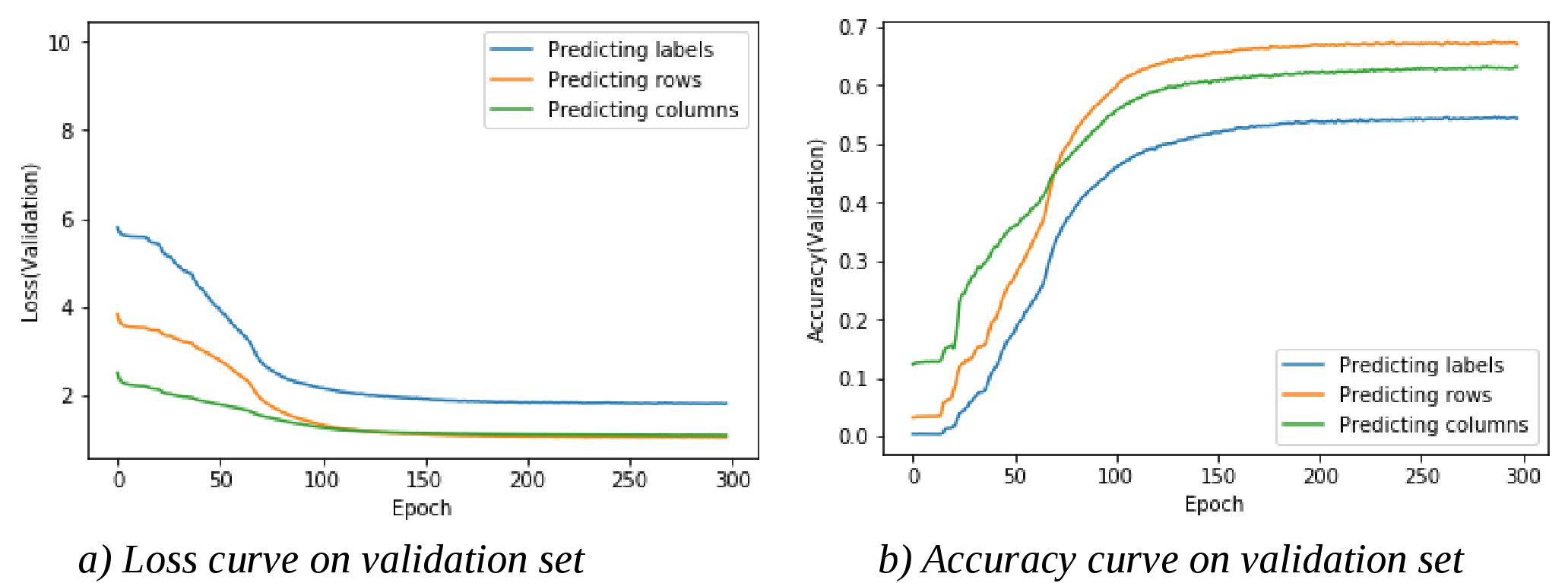}
%
% If no graphics program available, insert a blank space i.e. use
%\picplace{5cm}{2cm} % Give the correct figure height and width in cm
%
\caption{Examining the learning behavior of the auxiliary tasks (\textalpha1=1 \textalpha2=0.35 \textalpha3=0.65)}
\label{MesaySamuel:8}       % Give a unique label
\end{figure}

\noindent From Figure ~\ref{MesaySamuel:7}, it was observed that the model with alpha values (\textalpha1=1 \textalpha2=0.35 \textalpha3=0.65) has performed best that enabled the fastest convergence. This has implied the significance of learning the columns on supporting the main task. Accordingly a closer investigation was made to examine how auxiliary tasks (predicting the rows and columns of the alphabet) have been learned together with the main task (predicting the individual alphabet labels). This was shown in Figure ~\ref{MesaySamuel:8}.

Figure ~\ref{MesaySamuel:8}-a shows how the loss drops for all the three tasks during the multi-task learning on validation set. Accordingly, the model predicted better for the auxiliary tasks. The loss dropped from 3.83 to 1.05 and from 2.51 to 1.09 for predicting the rows and columns respectively. Similarly, 54.35\%, 67.13\%, and 63.16\% accuracies are attained in predicting labels, rows, and columns respectively. Even though the model learned quickly for predicting the columns than rows, the rows achieved better performance on the latter epochs. This might have occurred due to over-fitting. However, the better prediction of the columns corresponds to the nature of alphabets under the same column exhibit a similar structuring over the base characters as shown in Figure ~\ref{MesaySamuel:1}. Finally, 54.92\%, 68.09\%, and 65.26\% accuracies are achieved on test set for predicting the labels, rows, and columns respectively. 

Apart from getting the improvement from the relationship of characters as placed in rows and columns, it was observed that predicting these auxiliary tasks are easier. Hence, these results will open a way to use these relationships for predicting labels and also predicting Amharic words. 

\section{Conclusion and Future Works}
\label{MesaySamuelsec:5}

In this study Amharic handwritten character recognition was addressed using a convolutional neural network. Without the need for hand crafted feature extraction it was observed that one can achieve a reasonable recognition result. More importantly, the relationship among the characters as placed in the Amharic alphabet has opened a way for multi-task learning. The result of the study demonstrated the relevance of these auxiliary tasks in supporting the recognition accuracy. Particularly, the trained model performed better at predicting the rows and columns of the alphabets. Hence, getting the advantage of the rows and columns, Amharic word predictions will be investigated in the future works. A further investigation on the scalability of the proposed technique for multi-script recognition will also be explored in the future works. In this study the limitation of unique handwritten dataset affected the performance of the models. Hence, a standard real dataset will be developed which can be used for related machine learning experiments. 

Finally, the authors would like to acknowledge Yaregal Assabie and Josef Bigun for providing the Ethiopic handwriting dataset. 

\bibliographystyle{natdin}  
\bibliography{MesaySamuel}

\end{document}